\documentclass[11pt, a4paper]{article}
\usepackage[utf8]{inputenc}
\usepackage{authblk}
\usepackage[numbers]{natbib}
\usepackage{graphicx}
\usepackage{hyperref}
\usepackage{cleveref}
\usepackage{enumitem}
\usepackage{siunitx}
\usepackage{booktabs}
\usepackage{subcaption}
\usepackage{numprint}
\usepackage{microtype} 
\usepackage{xspace}

\title{Transfer Learning for Olfactory Object Detection}
\author[1]{Mathias Zinnen}
\author[1]{Prathmesh Madhu}
\author[2]{Peter Bell}
\author[1]{Andreas Maier}
\author[1]{Vincent Christlein}
\affil[1]{\small Pattern Recognition Lab, Friedrich-Alexander-Universität Erlangen-Nürnberg (FAU), 91058 Erlangen, Germany}
\affil[2]{\small Germanistik und Kunstwissenschaften (Fb09), Philipps-Universität Marburg, 35032 Marburg, Germany}

\date{}

\newcommand{\ie}{i.\,e.,\xspace}
\newcommand{\eg}{e.\,g.,\xspace}

\begin{document}
\maketitle

\section{Introduction}
Smells are an important, yet overlooked part of cultural heritage \cite{bembibre2017smell}.
The Odeuropa project\footnote{www.odeuropa.eu} analyzes large amounts of visual and textual corpora to investigate the cultural dimensions of smell in 16th -- 20th century Europe. The study of pictorial representations bears a specific challenge:  
the substrate of smell is usually invisible \cite{marx2021}.

Object detection is a well-researched computer vision technique, and so we start with the recognition of objects, which may then serve as a basis for the indirect recognition of more complex, and possibly more meaningful, smell references such as gestures, spaces, or iconographic allusions \cite{zinnen2021see}.
\begin{figure}
    \centering
    \includegraphics[width=\textwidth]{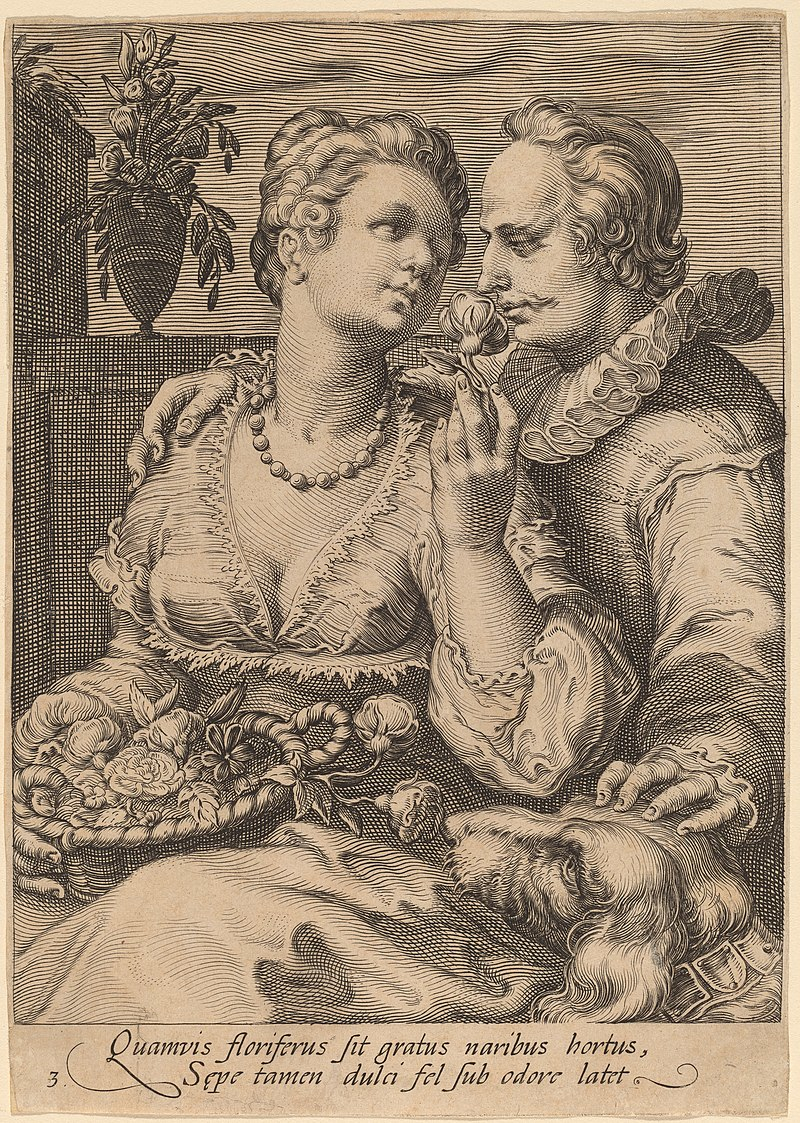}
    \caption{\textit{Smell}. The Five Senses. 1558 -- 1617. Jan Pietersz Saenredam. National Gallery of Art. Public Domain}
    \label{fig:sniff}
\end{figure}

However, the detection of olfactory objects in historical artworks is a challenging task. 
The visual representation of objects differs significantly between artworks and photographs \cite{hall2015cross}. Since state-of-the-art object detection algorithms are trained and evaluated on large-scale photographic datasets such as ImageNet \cite{russakovsky2015imagenet}, MS COCO \cite{lin2014microsoft}, or OpenImages \cite{kuznetsova2020open}, their performance drops significantly when applied to artistic data. 
This \emph{domain gap} between standard object detection datasets and artistic imagery can be mitigated by training directly on artworks, either by using existing datasets or by creating an annotated dataset for the target domain. 
Another challenge is the mismatch between object categories present in modern datasets and historical olfactory objects, caused by historical diachrony on the one hand \cite{marinescu2020improving}, and the particularity of some smell-relevant objects on the other\cite{zinnen2021see}, \cite{ehrich2020}.

\section{Methodology}

\begin{figure}[t]
    \centering
    \includegraphics{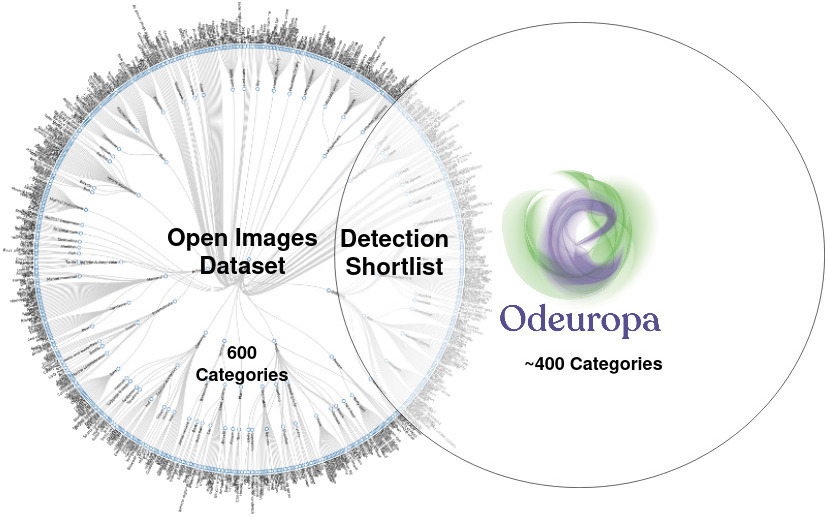}
    \caption{Category overlap between Odeuropa \& OpenImages categories}
    \label{fig:venn}
\end{figure}

To overcome the domain gap and category mismatch between our application and the existing datasets, we apply transfer learning -- a training strategy where machine learning algorithms are pre-trained in one domain and then fine-tuned in another, greatly decreasing the amount of required training data in the target domain (\cite{pan2009survey},\cite{zhuang2020comprehensive},\cite{madhu2020enhancing}].

We are continuously collecting and annotating artworks with possible olfactory relevance from multiple museum collections.
Based on these, we created a dataset of olfactory artworks containing \numprint{16728} annotations on \numprint{2229} artworks. 
From this full set of annotations, we created a test set of \numprint{3416} annotations on \numprint{473} artworks, while the remaining data was used for training.

\begin{figure}[t]
    \centering
    \includegraphics[width=\textwidth]{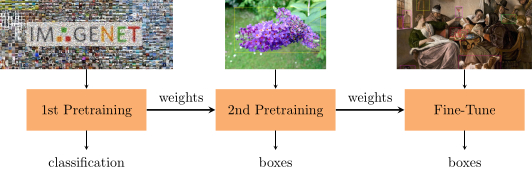}
    \caption{Transfer learning training strategy illustration. We start with a backbone pre-trained on ImageNet for classification, use this model to train an object detection system using different datasets. Finally, the object detection model is fine-tuned on the olfactory artworks dataset.}
    \label{fig:pipeline}
\end{figure}

A common transfer learning procedure is to use detection backbones that have been pre-trained on ImageNet and fine-tune them for object detection \cite{zhuang2020comprehensive}.
We expand this strategy by an additional pre-training step, where we train an ImageNet pre-trained object detection network \cite{ren2015faster} using different datasets. 
Finally, we fine-tune the resulting model using our olfactory artworks dataset (\cref{fig:pipeline}). 

For pre-training, we use three different datasets, deviating to varying amounts from our olfactory artworks dataset in terms of categories and style (\cref{tab:datasets}):
a) Same Categories, Different Styles - A subset of OpenImages (OI) containing only odor objects results in a complete category match (\cref{fig:venn}); however, since OpenImages contains only photographs, there is a considerable style difference. 
b) Different Categories, Same Styles - We apply two object detection datasets from the art domain, which are more similar in terms of style but contain different object categories, namely IconArt (IA) \cite{gonthier2018weakly} and PeopleArt (PA) \cite{westlake2016detecting}.

\begin{table}[t]
    \centering
    \caption{An overview of domain \& category similarity of the experiment datasets to our olfactory artworks}
    \begin{tabular}{lccc}
    \toprule
    Dataset & domain similarity & category similarity & \# categories \\
    \midrule
    OpenImages & low & complete match & 29\\
    IconArt    & high & medium & 10\\
    PeopleArt  & medium & low & 1\\
    \bottomrule
    \end{tabular}
    \label{tab:datasets}
\end{table}

\section{Results}

\begin{table}[t]
    \centering
    \caption{
    Evaluation of object detection performance.
    The best performing model pre-trained with OI achieves an improvement of 6.5\% pascal VOC mAP, and 3.4\% COCO mAP over the baseline method without intermediate training.
    We report the evaluation for each pre-training dataset,  averaged over five models, fine-tuned for 50 epochs on our olfactory artworks datasets.
    Best evaluation results are highlighted in bold.
    The merge of two datasets $D_1$ and $D_2$ is written as $D_1 \cup D_2$. 
    }
    \begin{tabular}{lcc}
    \toprule
        Pretraining Dataset &  Pascal mAP(\%) & COCO mAP(\%)\\
        \midrule
        None (Baseline) & 16.8($\pm$1.3) & 8.4($\pm$0.4) \\
        \midrule
        OI & \textbf{23.3($\pm$0.5)} & \textbf{11.8($\pm$0.4)} \\
        IA & 22.6($\pm$1.2) & 10.9($\pm$0.9) \\
        PA & 21.9($\pm$0.4) &  10.5($\pm$0.2) \\
        IA$\cup$OI & 21.8($\pm$0.1)  & 10.5($\pm$0.3)\\
        IA$\cup$PA & 22.0($\pm$0.8)  & 10.6($\pm$0.3)\\
        PA$\cup$OI & 22.6($\pm$0.3) & 10.8($\pm$0.2) \\
        OI$\cup$IA$\cup$PA & 21.8($\pm$0.4) & 10.5($\pm$0.2) \\
        \bottomrule
    \end{tabular}
    \label{tab:results}
\end{table}

\begin{figure}
    \begin{subfigure}[t]{.32\textwidth}
        \includegraphics[width=\textwidth]{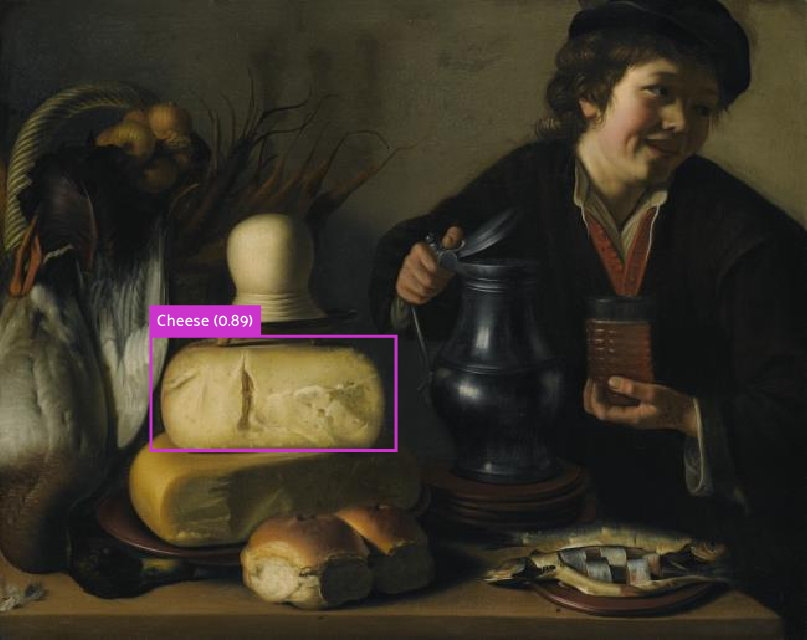}    
        \caption{No pretraining.}
    \end{subfigure}
    \begin{subfigure}[t]{.32\textwidth}
        \includegraphics[width=\textwidth]{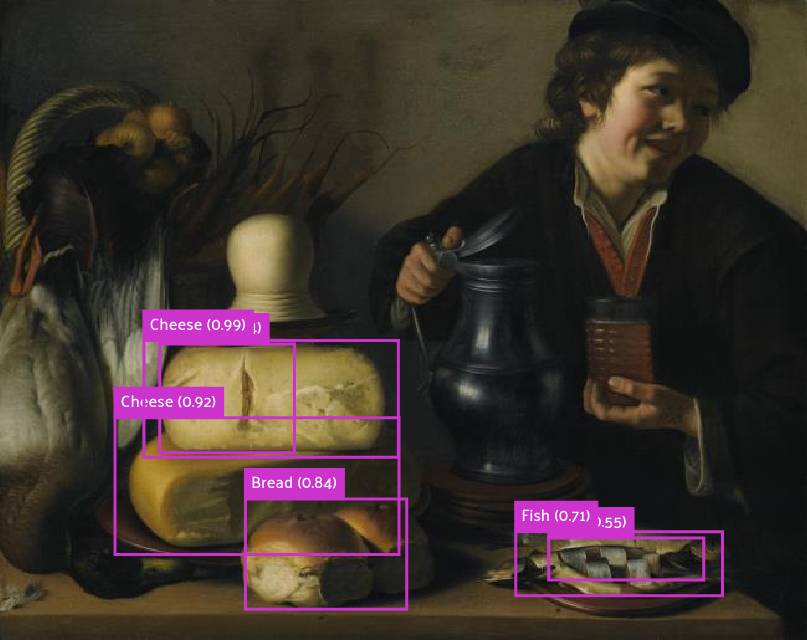}    
        \caption{PA pretraining.}
    \end{subfigure}
    \begin{subfigure}[t]{.32\textwidth}
        \includegraphics[width=\textwidth]{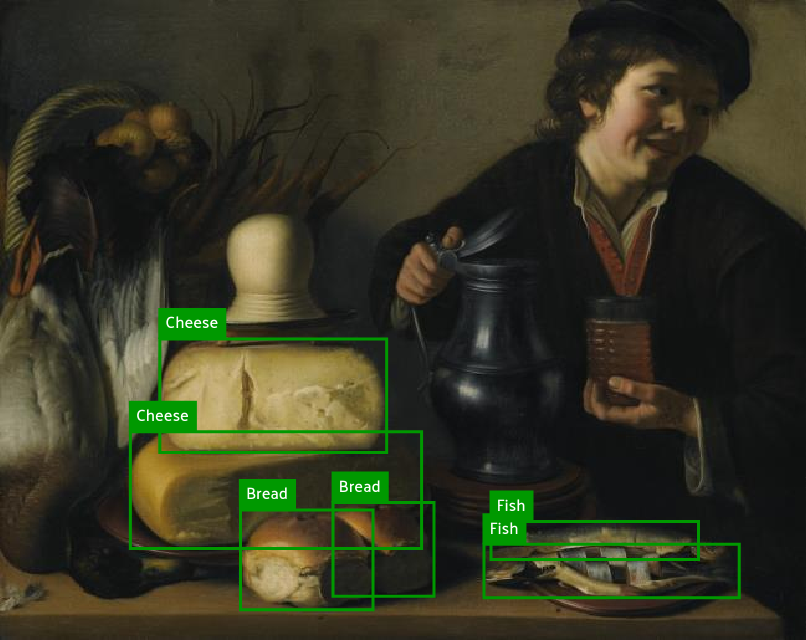}    
        \caption{Ground truth.}
    \end{subfigure}
    \caption{
    Exemplary object predictions for a detection model without intermediate training (a), with PeopleArt pretraining (b), and ground truth bounding boxes (c).
    Painting: \textit{Boy holding a pewter tankard, by a still life of a duck, cheeses, bread and a herring.} 1625 -- 1674. Gerard van Honthorst.
    RKD Digital Collection (\url{https://rkd.nl/explore/images/287165}). Public Domain.
    }
    
    \label{fig:preds}
\end{figure}

To ensure a fair comparison between the different pre-training datasets, we reduce each of the datasets to the same size, train three models, and select the best according to a fixed validation set for each dataset.
Additionally, we merge all three datasets, \ie combining OI, IA, and PA, using the union over their respective classes. 
The resulting models are then fine-tuned on the training set of the olfactory artworks dataset and evaluated on a separate test set. 
To mitigate random variations that can occur during the training process, we train five separate models for each experimental setting and report their average. 
Evaluation results are reported in pascal VOC (mAP 50 \cite{everingham2010pascal}) and COCO mAP (mAP 50:95:5 \cite{lin2014microsoft}), the two standard metrics to evaluate object detection models. 
We conduct two separate sets of experiments: 
In the first, we fine-tune the whole network, including the backbone, to assess the detection performance under realistic conditions (\cref{tab:results}).
We observe a performance increase for all used pre-training datasets, with an increase of 6.5\%/3,4\% boost in mAP 50 and COCO mAP, respectively, for the best performing pre-training scheme, which was achieved using the OI dataset.
The exemplary object predictions in \cref{fig:preds} show that adding an additional pre-training stage can increase the number of recognized objects.

\begin{table}[t]
    \centering
    \begin{tabular}{lcc}
    \toprule
        Pretraining Dataset &  mAP(\%) 50 & mAP(\%) 50:95\\
        \midrule
        None (Baseline) & 11.7($\pm$0.2) & 5.5($\pm$0.1) \\
        \midrule
        OI & \textbf{19.4($\pm$0.3)} & \textbf{9.5($\pm$0.1)} \\
        IA & 13.8($\pm$0.4) & 6.4($\pm$0.2) \\
        PA & 13.5($\pm$0.2) &  6.7($\pm$0.1) \\
        IA$\cup$OI & 16.0($\pm$0.3)  & 7.4($\pm$0.2)\\
        IA$\cup$PA & 14.6($\pm$1.0)  & 6.7($\pm$0.5)\\
        PA$\cup$OI & 15.8($\pm$0.7) & 7.3($\pm$0.4) \\
        OI$\cup$IA$\cup$PA & 16.4($\pm$0.6) & 7.6($\pm$0.2) \\
        \bottomrule
    \end{tabular}
    \caption{
        Evaluation of object detection performance for fine-tuning of the detection heads only.
        All pre-training schemes increase the detection performance, while pre-training with OI leads to the best results with an increase of 7.7\% mAP 50 or 4\% COCO mAP. 
    For every pre-training dataset, we report the evaluation averaged over five models, fine-tuned for 50 epochs on our olfactory artworks datasets each.
    Best evaluation results are marked in bold.
    The merge of two datasets $D_1$ and $D_2$ is written as $D_1 \cup D_2$. 
    }
    \label{tab:headonly}
\end{table}

In a second set of experiments, we train only the detection head while the backbone remains frozen, to compare the quality of the intermediate representations that have been learned using the different pre-training schemes (\cref{tab:headonly}).
While all pre-training schemes increase the performance, the relative increase for the OI dataset is remarkably higher.
This suggests that the style similarity between the IA and PA datasets and our target dataset is less important than we expected.
We can not yet conclude whether the superior performance of the OI dataset is due to the similarity in target categories.
It could also be caused by other properties of the dataset.
Further ablations, \eg varying the set of OI categories are needed to more precisely assess the impact of category similarity on the detection performance, which we plan to conduct in a follow-up study.
Interestingly, the performance of the merged datasets increases even in cases where OI is not part of the dataset merge.
Given that we did not apply a sophisticated merging strategy, the performance increase for training with merged datasets is encouraging. 
Developing strategies to improve the consistency of the merged dataset, \eg weak labeling of categories not present in the respective merge partners, represents another promising line of future research.

We conclude that including an additional stage of object-detection pre-training can lead to a considerable increase in detection performance.
While our experiments suggest that style similarities between pre-training and target dataset are less important than matching categories, further experiments are needed to verify this hypothesis. 

\section{Acknowledgements}
We gratefully acknowledge the support of NVIDIA Corporation with the donation of the two Quadro RTX 8000 used for this research. 
The paper has received funding by Odeuropa EU H2020 project under grant agreement No. 101004469.

\clearpage
\bibliographystyle{plain}
\bibliography{main}

\end{document}